\documentclass{article}

\PassOptionsToPackage{numbers,square,sort&compress}{natbib}

\usepackage[dblblindworkshop, final]{neurips_2025}
\workshoptitle{AI for Science}
\setcitestyle{square,numbers,sort&compress}

\usepackage[utf8]{inputenc} 
\usepackage[T1]{fontenc}    
\usepackage{hyperref}       
\usepackage{url}            
\usepackage{booktabs}       
\usepackage{amsfonts}       
\usepackage{nicefrac}       
\usepackage{microtype}      
\usepackage{xcolor}         

\usepackage[table]{xcolor}
\usepackage{booktabs}
\usepackage{array}
\usepackage{amsmath}
\usepackage{makecell}   
\usepackage{hyphenat}   

\usepackage[most]{tcolorbox}
\usepackage{fancyvrb}
\usepackage{adjustbox}

\usepackage[most]{tcolorbox}
\usepackage{enumitem}
\usepackage{lmodern}

\definecolor{ExplicitBG}{RGB}{255,235,238} 
\definecolor{PremiseBG}{RGB}{232,244,253}  
\definecolor{GapBG}{RGB}{255,243,224}      

\newcolumntype{Y}{>{\raggedright\arraybackslash}X}
\renewcommand{\arraystretch}{1.15}
\usepackage{listings}
\usepackage{xcolor}
\usepackage[T1]{fontenc}

\usepackage[numbers,sort&compress]{natbib}

\usepackage[T1]{fontenc}
\usepackage{lmodern}
\usepackage{microtype}

\usepackage{booktabs}
\usepackage{tabularx}
\usepackage{array}
\usepackage{ragged2e}
\usepackage{multirow}

\usepackage[table]{xcolor}
\usepackage[most]{tcolorbox}
\tcbset{
  highlight/.style={
    on line,
    boxrule=0pt,
    colframe=black!0,
    arc=2pt,
    left=1.5pt,right=1.5pt,top=0.5pt,bottom=0.5pt 
  }
}
\definecolor{amber}{RGB}{255,217,128}
\newcommand{\hlred}[1]{\tcbox[colback=red!18,  highlight]{#1}}
\newcommand{\hlblue}[1]{\tcbox[colback=blue!15, highlight]{#1}}
\newcommand{\hlamber}[1]{\tcbox[colback=amber!70,highlight]{#1}}

\usepackage{csquotes}
\newcolumntype{Y}{>{\RaggedRight\arraybackslash}X}
\newcolumntype{P}[1]{>{\RaggedRight\arraybackslash}p{#1}}
\renewcommand{\arraystretch}{1.18}

\newcommand{\best}[1]{\textbf{#1}}
\newcommand{\second}[1]{\underline{#1}}

\usepackage{booktabs,longtable}
\usepackage{multirow}
\usepackage{caption}

\title{GAPMAP: Mapping Scientific Knowledge Gaps in Biomedical Literature Using Large Language Models}

\author{%
  Nourah M.~Salem \\
  University of Colorado, Anschutz\\
  University of Chicago \\
  \texttt{nourah.salem@cuanschutz.edu}
  \And
  Elizabeth White \\
  University of Chicago \\
  \texttt{elizabeth.white@bsd.uchicago.edu}\\
  \And
  Michael Bada \\
  University of Chicago \\
  \texttt{michael.bada@bsd.uchicago.edu} \\
  \And
  Lawrence Hunter \\
  University of Chicago \\
  \texttt{lehunter@uchicago.edu}
} 

\newcommand{\TABIcal}{%
  \ensuremath{%
    \mathcal{T}\kern-0.18em
    \mathcal{A}\kern-0.04em
    \mathcal{B}\kern-0.04em
    \mathcal{I}}}

\begin{document}

\maketitle

\begin{abstract}
  Scientific progress is driven by the deliberate articulation of what remains unknown. This study investigates the ability of large language models (LLMs) to identify research knowledge gaps in the biomedical literature. We define two categories of knowledge gaps: explicit gaps, clear declarations of missing knowledge; and implicit gaps, context-inferred missing knowledge. While prior work has focused mainly on explicit gap detection, we extend this line of research by addressing the novel task of inferring implicit gaps. We conducted two experiments on almost 1500 documents across four datasets, including a manually annotated corpus of biomedical articles. We benchmarked both closed-weight models (from OpenAI) and open-weight models (Llama and Gemma 2) under paragraph-level and full-paper settings. To address the reasoning of implicit gaps inference, we introduce \textbf{\small TABI}, a Toulmin-Abductive Bucketed Inference scheme that structures reasoning and buckets inferred conclusion candidates for validation. Our results highlight the robust capability of LLMs in identifying both explicit and implicit knowledge gaps. This is true for both open- and closed-weight models, with larger variants often performing better. This suggests a strong ability of LLMs for systematically identifying candidate knowledge gaps, which can support early-stage research formulation, policymakers, and funding decisions. We also report observed failure modes and outline directions for robust deployment, including domain adaptation, human-in-the-loop verification, and benchmarking across open- and closed-weight models.
  
\end{abstract}

\section{Introduction}

The rapid growth of scientific publications has made it difficult for researchers to stay aware of unanswered questions and knowledge gaps (known unknowns) in their fields \cite{boguslav2023creating}.
Identifying such gaps is critical for prioritizing new studies and directing funding toward consequential open problems \cite{boguslav2023creating, abd2024machine}.
Traditionally, researchers surface gaps via exhaustive literature reviews and expert judgment, a process that is time-consuming and hard to scale to the current volume of literature.
This motivates automated methods that extract knowledge gaps directly from scientific text to accelerate discovery.

Knowledge gaps exist in different forms and span various types in the literature. We build on prior work that organizes author-signaled gap statements into taxonomies of gap types (e.g., Levels of Evidence, Barriers, Future Opportunities, Anomalous findings, and Research Aims) \cite{boguslav2023creating}.
On top of this, we introduce another layer that distinguishes \textbf{explicit} from \textbf{implicit} instances within the same semantic categories.
An instance is explicit when the gap is directly signaled by high uncertainty lexical cues (e.g., “unknown,” “further research is needed”), or low uncertainty hedging words (e.g., "May lead to," "could cause") or negation in some cases, typically recoverable at the sentence or short-paragraph level (e.g., “It remains unknown whether long-term GLP-1 agonist therapy improves renal outcomes in non-diabetic CKD”).
An instance is implicit when the gap is not stated directly by the authors of the research papers, but can be inferred from section-level context, such gaps include: (i) a chain of claims containing a missing link; (ii) a generalization gap where claims are developed under a limited scope but out-of-scope applicability is questionable; or (iii) conflicting findings presented without reconciliation or an experiment to resolve them.
This explicit–implicit overlay preserves the meaning of existing gap categories while extending coverage to unstated, context-dependent gaps. Table 1 provides examples of explicit and implicit knowledge gaps.

\paragraph{Our main contributions are as follows:}

\begin{itemize}
\item \textbf{Benchmarking open- and closed-weight LLMs.} We introduce a context-aware evaluation approach, which spans two experimental designs and four public datasets (two per gap type), and report reproducible baseline results, along with their insights.

\item \textbf{Implicit knowledge gap inference.} To our knowledge, this is the first systematic evaluation of LLMs on inferring unstated knowledge gaps at the section/document scale, requiring discourse-level reasoning beyond lexical cues.

\item \textbf{TABI: Toulmin-Abductive Bucketed Inference.}

We cast implicit-gap detection as abductive Natural Language Inference (NLI) with generation. The model follow Toulmin Argument Model \cite{karbach1987using} to generate a \emph{Claim} justified by \emph{Grounds$\to$Warrant}. Claims are bucketed into \emph{more} vs.\ \emph{least} probable conclusions. This approach enables overlap-based scoring against gold conclusions and premises and potentially tracking the sources of errors in making these conclusions.

\end{itemize}

\begin{table}[t]
\footnotesize
\caption{Examples of knowledge-gap statements in biomedical literature. Cues are highlighted: explicit lexical cues in red, implicit premises in blue, and the inferred gap in amber.}
\begin{tabularx}{\textwidth}{P{2.0cm} Y P{3.8cm}}
\toprule
\textbf{Type} & \textbf{Example with highlighted cues} & \textbf{Cue(s)} \\
\midrule
\textbf{Explicit}
  & \enquote{\hlred{X remains unknown}.}
  & \enquote{remains unknown} \\
\addlinespace[0.15em]
\textbf{}
  & \enquote{To date, \hlred{no randomized controlled trial} has evaluated therapy Y in condition Z.}
  & \enquote{No RCT} \\
\midrule
\textbf{Explicit}
  & \enquote{\hlred{X remains unknown}.}
  & \enquote{remains unknown} \\
\addlinespace[0.15em]
\textbf{}
  & \enquote{To date, \hlred{no randomized controlled trial} has evaluated therapy Y in condition Z.}
  & \enquote{No RCT} \\
\midrule
\textbf{Implicit}
  & \begin{tabular}[t]{@{}l@{}}
      \textbf{Grounds:}\\[0.25em]
      \hlblue{Compound E improves biomarker F in mice.}\\
      \hlblue{Biomarker F correlates poorly with clinical outcomes in humans.}\\[0.25em]
      $\Rightarrow$ \textbf{Inferred gap (Claim):}\\ \hlamber{It is unknown whether E improves patient outcomes.}
    \end{tabular}
  & Outcome mismatch; translational gap \\
\bottomrule
\end{tabularx}
\medskip
\emph{Notes.} Examples are illustrative. Explicit gaps use lexical markers; implicit gaps arise when jointly considering premises that imply a missing piece of evidence.
\end{table}

\section{Related Work}

Prior literature started with related tasks, such as labeling uncertainty/hedging, negation and speculation in clinical NLP. One of the first significant advancements was introduced by Hyland \cite{hyland1996writing}, who studied hedging linguistic forms in relation to their pragmatic use in cell and molecular biology articles. Hyland highlighted how researchers hedge their language to qualify levels of uncertainty in their findings. Similarly, Goujon \cite{goujon2009uncertainty} explored the use of linguistic cues to classify uncertainty levels in sentences, using predefined linguistic patterns to identify expressions of uncertainty (e.g., "doubtful," "Could be"). While these approaches can be highly effective for well-defined and structured phrases, their reliance on predefined linguistic patterns limits their scope.

In another early study, Chapman, \emph{et al} \cite{chapman2001simple} developed the rule-based algorithm “NegEx” to detect uncertainty by identifying negated findings in medical discharge summaries. This method was effective in its clinical setting, though it struggled with more complex text analyses due to its lack of semantic understanding. 

Building on this, Chen, \emph{et al} \cite{chen2018scalable} expanded \cite{hyland1996writing}’s work by adding 61 cue words to the original seed list and using the Word2Vec technique to identify semantically related words. Although Word2Vec captured semantic similarity, it did not fully address the contextual levels of uncertainty that can arise in scientific text.

Recent efforts have shifted from purely linguistic patterns to machine learning models for ignorance extraction, as these models can capture not just the explicit uncertainty cues, but also subtleties like implications, or patterns of omission which might reflect knowledge gaps around certain ideas. Boguslav, \emph{et al} \cite{boguslav2023creating, boguslav2021identifying} developed a methodology to categorize and identify explicit unknowns in scientific literature by tuning Conditional Random Fields (CRF) \cite{sutton2012introduction} and BERT models \cite{vaswani2017attention}. They applied these models to classify parsed statements from prenatal nutrition articles as either ignorance statements or not, followed by multi-class classification to categorize these knowledge gap statements by type. This approach demonstrated strong performance with BERT and its variants (e.g., BioBERT), though generalizability and scalability remain challenges due to the size of the BERT models. 

More recently, Bibal \emph{et al} \cite{bibal2024recsoi} launched "RecSOI," a system recommending research knowledge gaps by based on researchers interests. Their work touched on the use of large language models (LLMs), specifically GPT-3.5 and GPT-4, to recognize ignorance in text as a recommender system. However, they did not provide a quantitative analysis of their experiments with LLMs. Our project builds on these efforts by using large language models to extract the explicit knowledge gap statements and infer the implicit ones, with the goal of advancing the understanding of scientific knowledge gaps.

\section{Datasets and Models}
\vspace{-3mm}
We used four datasets to study explicit and implicit knowledge gaps across scientific literature, with varying levels of structure and scale. As this is current ongoing project, one of our ultimate goals is to test and enrich these benchmarking datasets, specially the implicit gap ones.
\vspace{-1.5mm}
\subsection{Explicit Knowledge Gaps}
\vspace{-1.5mm}
\paragraph{IPBES \cite{ipbes_knowledge_gaps_page5}}  
A structured collection of 286 short paragraphs from biodiversity assessments, containing 657 labeled knowledge gap statements. Designed to guide policy-relevant research in biodiversity and ecosystems saving.
\vspace{-2.5mm}
\paragraph{Scientific Challenges \& Directions \cite{lahav2022search}}  
Contains $\sim$2,900 annotated sentences from 1,786 COVID-19 research papers, each labeled binary as carrying scientific challenges and/or research directions or not. Sentences are embedded within full sections, which in some cases exceed 8K tokens size, posing challenges for certain LLMs with smaller attention windows.
\vspace{-1.5mm}
\subsection{Implicit Knowledge Gaps}
\vspace{-1.5mm}
\paragraph{Paragraph-Level Manual Implicit-Gap Dataset}  
A small benchmark of 212 biomedical paragraphs from 137 PubMed articles, manually annotated by our lab for implicit future-direction gap statements. Each paragraph consists of a set of premises and  masked conclusion statements that must inferred by the LLM. Dataset and code available on GitHub.\footnote{\url{https://github.com/UCDenver-ccp/GAPMAP}}
\vspace{-2.5mm}
\paragraph{Full-Text Comprehension Pilot Dataset}  
Aimed at evaluating the feasibility of using LLMs to analyze full research manuscripts for knowledge gaps and future directions. The dataset for this pilot experiment includes 24 full-text articles from 19 scientific domains (e.g., Immunology, Astrophysics, Materials Science), processed using GPT-4o interface.

\subsection{Models}
\noindent We used a group of LLMs from different families to test their capacity, modality, and efficiency. From Meta, we selected Llama-3.3-70B-Instruct \cite{llama33}; Llama-3.1-8B-Instruct as an efficient text-only baseline \cite{llama31}; Llama-4~Scout~17B, a Mixture-of-Experts model (17B active parameters, 16 experts) \cite{llama4scout}; gemma-2-9b-it  from Google  \cite{gemma2}; and for the closed models, we tested OpenAI’s GPT-5, GPT-4o, and GPT-4o~mini \cite{gpt5,gpt4o,gpt4omini}. This combination allow us to probe method performance across scales (8B–70B+), modalities (text and multimodal), and deployment constraints (latency/cost), strengthening the generality of our findings.

\begin{table}[t]
\centering
\small
\caption{Datasets used in this study for explicit and implicit knowledge gaps.}
\label{tab:data-overview}
\begin{tabularx}{\linewidth}{@{}p{1.6cm} Y p{3.2cm} p{3.6cm} p{2.4cm}@{}}
\toprule
\textbf{Category} & \textbf{Source} & \textbf{Unit} & \textbf{Scope / Domain} & \textbf{Size} \\
\midrule
Explicit &
IPBES  &
Paragraph-Level &
Biodiversity \& Ecosystem &
657 statements from 286 paragraphs \\

 &
Scientific Challenges \& Directions &
Section-Level &
COVID-19 &
2,894 statements from 1,786 Studies\\
\midrule
Implicit &
Full-Text Pilot Dataset &
Full-Paper-Level &
STEM Field &
23 Studies \\

 &
Manual implicit-gap corpus&
Paragraph-Level &
Biomedical Field&
212 Paragraphs from 137 articles \\
\bottomrule
\end{tabularx}
\end{table}

\section{Experimental Setup}\label{sec:exp-explicit}
\subsection{Explicit Knowledge Gaps}

Here, we evaluate the ability of large language models (LLMs) to identify explicit scientific knowledge gaps within scientific literature, as statements of uncertainty, limitations, contradictions, or missing evidence from structured scientific text inputs.

Let $\mathcal{C}$ denote a corpus composed of scientific sections, where each section $\mathcal{S}_i \in \mathcal{C}$ consists of a sequence of paragraphs:
$$\mathcal{S}_i = \{p_1, p_2, \ldots, p_n\}$$
Each paragraph $p_j$ is in turn composed of a sequence of sentences:
$$p_j = \{s_{j1}, s_{j2}, \ldots, s_{jm}\}$$

Given the IPBES dataset, the LLMs were fed with a sequence of $p_j$, which has average word count of 257, yet some of the paragraphs exceeded the 2500 words. This was tolerable with the set of models we're testing, however, when working with the COVID-19 dataset, Sections from the PubMed articles often exceeded the 8K token limit, which led us to using the chunking strategy, with attention to small models. We also had to filter out the non-gap statements and exclude the ones that reflect general non research knowledge gaps (e.g., \emph{\second{"SARS-CoV-2 has been declared a pandemic that causes COVID-19".}})  

\subsection{Chunking Strategy}
Each section $\mathcal{S}_i$ is segmented into chunks $\mathcal{T}_k$ using Stanza parser \cite{qi2020stanza} for all chunking experiments to maintain the same input size across models, such that:
$$
|\mathcal{T}_k| \leq 1000 \text{ words}
$$
Chunk boundaries are aligned at sentence boundaries to preserve semantic coherence. If a sentence $s_{kl}$ would cause the chunk to exceed 1000 words, it is deferred to the next chunk $\mathcal{T}_{k+1}$.

\subsection{Implicit Knowledge Gap Identification}
\label{sec:implicit_gap_design}

\subsubsection{Experiment 1: Paragraph-Level Inference with Manual Annotations}
Implicit gap statements are defeasible. They arise as the best explanation of observed premises rather than as explicit claims. In our task, we believe the representation must (i) capture defeasible inference, (ii) remain lightweight for short models inferences, (iii) follow the a certain pattern to help validating them, and (iv) transfer across domains and granularity (paragraph and document level). A minimal Toulmin scheme,\emph{Claim–Grounds–Warrant (CGW)}, meets these constraints, where the Warrant records the general justification that signals the gap from the cited Grounds, yielding interpretable outputs and enabling automatic consistency checks.

Here, $\mathcal{D}_{\text{manual}}=\{p_1,\dots,p_N\}$ is $N$ expert-annotated paragraphs list, each with one gold implicit gap $\gamma_i$, refering to a required future direction. We provide $p_i$ to the model, masking the statements of future directions claims that are at the end of these $p_i$ and ask the model to identify the following:
\vspace{-2.5mm}
\begin{itemize}
    \item \textbf{Claim} : the implied gap.
    \item \textbf{Grounds}: the evidence span(s) that support the Claim. 
    \item \textbf{Warrant}: A single sentence that reasons the Grounds to Claim. This is used as coherence and entailment sanity check (e.g., simple NLI or rule-based patterns), which aids human validation by exposing the inferential leap.
    \item \textbf{Bucket}: A binary classification of the model confidence in its inference. This is used as a calibration check: Does the \texttt{more\_probable} bucket contain a higher fraction of correct matches?
\end{itemize}

A paragraph is counted as a success if any predicted claim matches the gold gap. We also report where a match occurs (\texttt{more\_probable} vs. \texttt{least\_probable}) and the distribution of predicted categories to compare between what the LLM suggests as relevant and the ground truth actually indicates.

\subsubsection{Experiment 2: Document-Level Inference from Full Manuscripts}

As this is a pilot experiment aimed at examining the potential use of current LLMs on full articles, we only assessed GPT-4o interface, a multi-modal LLM, to parse different input articles formats and handle long context inputs. In this experiment, the model is exposed to each paper $\mathcal{P}_i$ as a full-document. Unlike selective input in the previous strategies, this experiment evaluates the model’s ability to reason holistically over the complete text.

Each full paper is passed to GPT-4o to generate structured outputs of pairs identifying (1) Implied knowledge gaps supported by textual evidence and (2) Suggested future direction.

In separate surveys, 18 corresponding authors of each paper reviewed the model's inferred gaps and proposed directions, and indicated whether they agreed or disagreed with each. In cases of disagreement, authors were asked to briefly justify their assessment (e.g., irrelevance, misinterpretation, outdated issue). This feedback was used to assess the quality and feasibility of full-document inference as a method for automated gap discovery.

\section{Evaluation}
\noindent We evaluate the IPBES explicit gap extractions against its ground truth using one-to-one matching based on ROUGE-L \textbf{F1} similarity (with stemming). Predictions are counted as true positives if the score exceeds a threshold of 0.55. Unmatched predictions are false positives, and unmatched gold sentences are false negatives. Exact string matches are also recorded for reference, but are not required for correctness. Aggregate precision, recall, and F1 are then reported across documents in Table 3. 

For subsequent evaluations (explicit extractions from the COVID-19 dataset and implicit gap inference), the LLMs either extracted extra valid unannotated explicit statements, and concluded more potentially valid claims than found in the gold standard dataset. For the COVID-19 experiment, we avoided penalizing plausible inferences. We validated these predictions against the domain-specific ignorance-cues dictionary of Boguslav, \emph{et al}.~\cite{boguslav2023creating} to confirm that each prediction at least carries one knowledge gap cue. Then, we adopt accuracy as the primary metric, reflecting the proportion of correctly identified gap statements. 

For the implicit-gap experiments, we computed accuracy by validating the Pairs: (Claim and gold) \&  (Warrant and Premises). Specifically, a prediction was marked correct if the bi-directional entailment probability using RoBERTa large model \cite{cohen2021exploring} exceeded a threshold of 0.4 to one of the claims. This techniques directly captures whether a claim is logically supported by its context rather than relying solely on the gold.

\section{Results and Discussion}

\begin{table}[t]
  \centering
  \small
  \caption{IPBES results under two context settings. Best F1 per block in \best{bold}, second best \second{underlined}.}
  \label{tab:ipbes-main}
  \setlength{\tabcolsep}{5pt}
  \begin{tabular}{lccc@{\hskip 8pt}ccc}
    \toprule
    & \multicolumn{3}{c}{No context window limit} & \multicolumn{3}{c}{1000-word chunks} \\
    Model & P & R & F1 & P & R & F1 \\
    \midrule
    \texttt{GPT-5}     & 0.7483 & 0.8476 & \second{0.7949} & 0.7376 & 0.8613 & 0.7947 \\
    \texttt{gpt-4o}    & 0.8445 & 0.4223 & 0.5630          & 0.8653 & 0.6463 & 0.7400 \\
    \texttt{GPT-4o mini} & 0.8013 & 0.7805 & 0.7907          & 0.7971 & 0.8323 & \best{0.8143} \\
    \texttt{Gemma-2-9B}  & 0.8091 & 0.5945 & 0.6854          & 0.7850 & 0.7000 & 0.7401 \\
    \texttt{Llama-3.3-70B}  & 0.8063 & 0.8567 & \best{0.8307}   & 0.7839 & 0.8460 & \second{0.8138} \\
    \texttt{Llama-4-17B}  & 0.7953 & 0.7568 & 0.7756          & 0.7583 & 0.7698 & 0.7640 \\
    \texttt{Llama-3.1-8B}  & 0.7711 & 0.5290 & 0.6275          & 0.7825 & 0.5595 & 0.6524 \\
    \bottomrule
  \end{tabular}
\end{table}
\paragraph{Explicit Gap Extraction with IPBES} 
The LLMs perform strongly in both full and chunked paragraph inputs, but with clear differences. 
Without a context limit, Llama-3.3-70B showed the best F1 driven by the highest recall, while GPT-5 and GPT-4o Mini are close behind. GPT-4o is the most conservative model, indicating a high-precision/low-recall operating point. 
When chunking the input into 1K-word chunks, GPT-4o Mini leads F1, narrowly ahead of Llama-3.3-70B, with GPT-5 remaining competitive. The top systems shift by $\approx$0.02 absolute F1 relative to the no-limit setting, showing that chunking preserves performance and can even raise recall for some models (notably GPT-4o Mini).

We conclude that open-weight models are competitive with closed-weight ones on lexically signaled gaps. Practically, Llama-3.3-70B and GPT-5 are preferable for broad coverage. Chunking is a safe preprocessing choice for explicit gaps and can improve recall without materially harming the overall F1 of smaller models such as Llama-3.1-8B.

\paragraph{Explicit Gap Extraction with COVID-19}
The COVID-19 dataset presents a more challenging setting for explicit knowledge-gap detection compared to IPBES. Unlike the IPBES annotations, which often align with clear lexical cues of uncertainty, the COVID corpus identifies only one gap statement per section, despite many sections containing multiple latent gaps. Furthermore, authors frequently relied on numeric contrasts, anomalies, or novel findings, rather than explicit uncertainty markers to convey gaps. As a result, LLMs often produced more candidate gap statements than the ground truth provided, yet, they are focused on the presence of uncertainty cues. 

Performance in Table 4 reflects these annotation and style differences. GPT-5 achieved the highest accuracy, though absolute values declined compared to IPBES. Mid-sized open-weight models (Llama-70B, Llama-17B) showed moderate alignment with annotated cues, while smaller models struggled beyond lexical markers. These results underscore the robustness of large-scale LLMs in contexts with fewer explicit signals and the limitations of benchmarks in capturing author-implied gaps. They also suggest that evaluation must consider not only lexical cues, but also numeric and contrastive reasoning patterns that signal gaps in biomedical discourse.


\begin{table}[t]
\centering
\small
\caption{Accuracy on two setups (\(n=973\) each). Highest accuracy per setup is in \textbf{bold}, second best \second{underlined}.}
\label{tab:acc-two-setups}
\begin{tabular}{@{}l rr rr@{}}
\toprule
& \multicolumn{2}{c}{\textbf{1000-word chunks}} & \multicolumn{2}{c}{\textbf{No context limit}} \\
\cmidrule(lr){2-3} \cmidrule(l){4-5}
\textbf{Model} & \textbf{Correct} & \textbf{Acc. (\%)} & \textbf{Correct} & \textbf{Acc. (\%)} \\
\midrule
\textbf{All Models}     & 782 & \textbf{80.37} & 730 & \textbf{75.03} \\
\second{GPT-5}     & 618 & \second{63.51} & 590 & \second{60.64} \\
Llama-3.3-70B & 406 & 41.73 & 385 & 39.57 \\
Llama-4-17B        & 362 & 37.20 & 378 & 38.85 \\
GPT-4o Mini & 301 & 30.94 & 419 & 43.06 \\
GPT-4o             & 271 & 27.85 & 252 & 25.90 \\
Gemma 2-9B         & 167 & 17.16 & 169 & 17.37 \\
Llama-3.1-8B       & 140 & 14.39 & 169 & 17.37 \\
\bottomrule
\end{tabular}
\end{table}

The three plots in Figure 1 and 2 summarize the additional explicit gap statements extracted from the COVID-19 dataset, and passed post-hoc validation via the ignorance-cues dictionary of Boguslav \emph{et al}.~\cite{boguslav2023creating}. 
The  Venn diagram in Figure 1 is showing how the four top models’ predictions overlap. About 1.14k items are shared by all four, but each model, especially GPT-5, also has many unique items, suggesting complement when using a mixture of these LLMs for certain tasks. 

\begin{figure}[t]
\centering
\begin{minipage}[t]{0.58\linewidth}
  \centering
  \includegraphics[width=\linewidth]{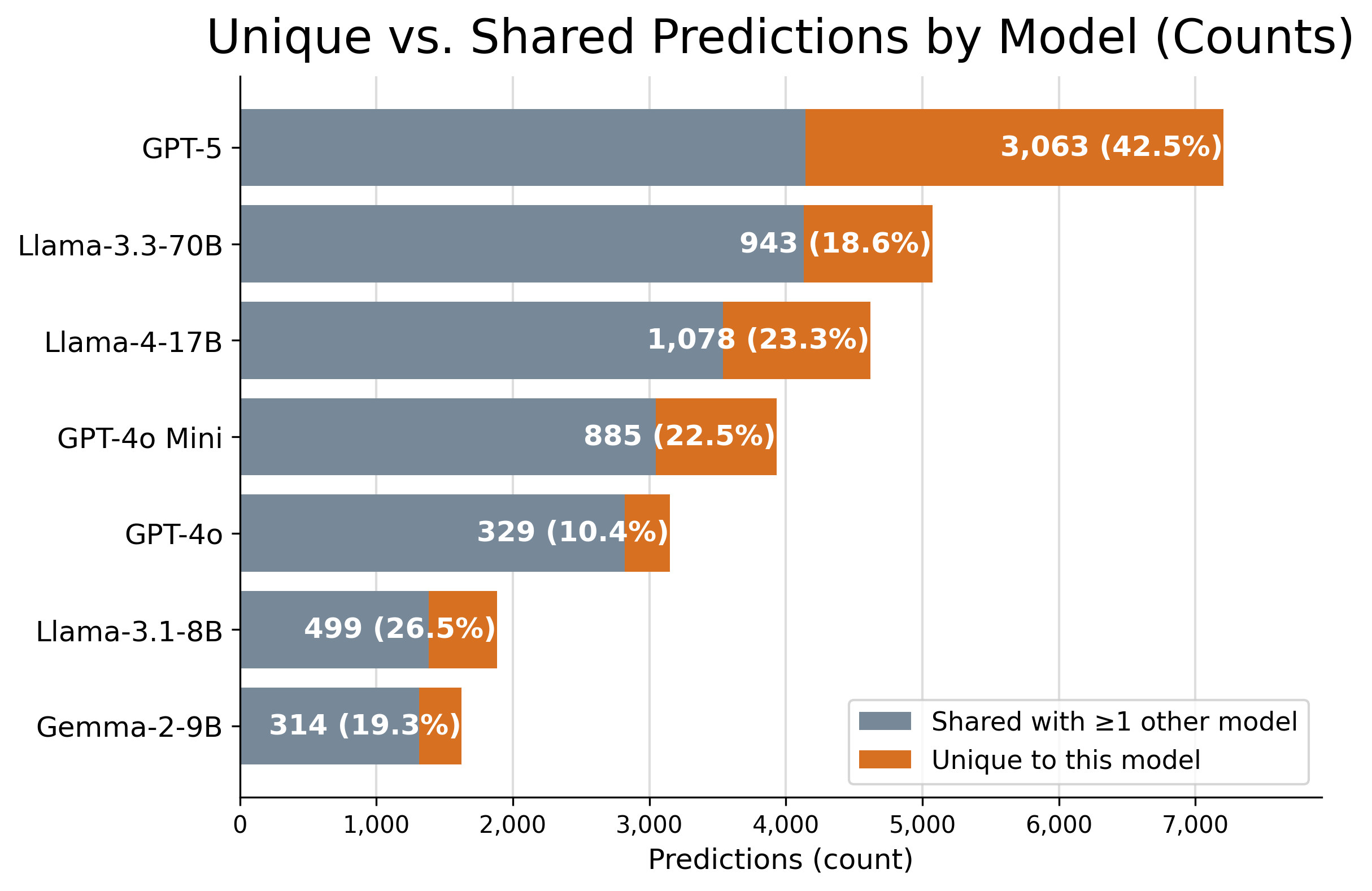}
\end{minipage}\hfill
\begin{minipage}[t]{0.35\linewidth}
  \centering
  \includegraphics[width=\linewidth]{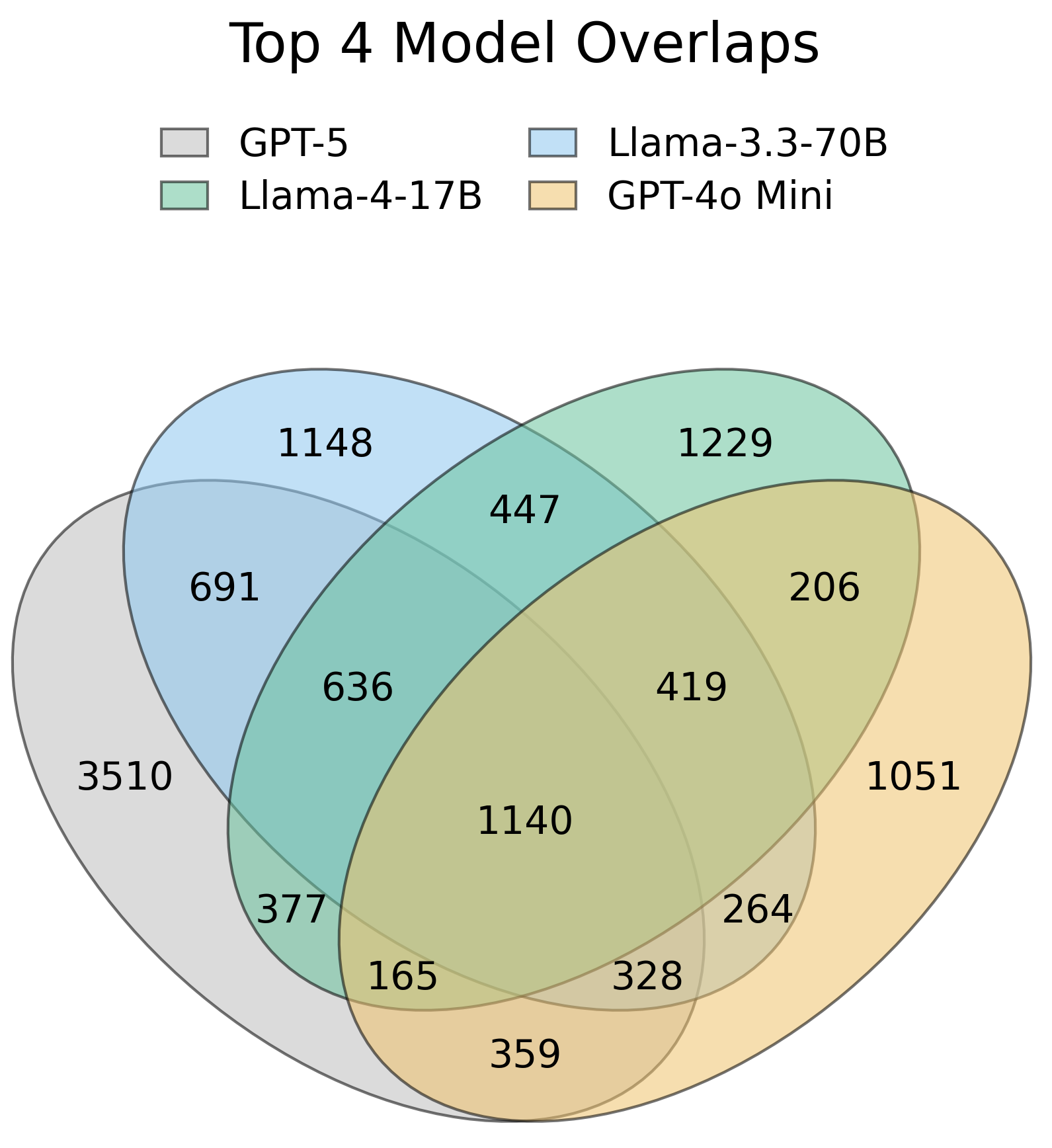}
\end{minipage}
\caption{\textbf{Model agreement and complementarity.} \textbf{Left:} For each model, predictions unique to that model (orange) vs.\ those shared with $\geq$1 other model. \textbf{Right:} Overlap among the top four models. Numbers denote instance counts per region.}
\label{fig:model-predictions}
\end{figure}

The left side diagram on Figures~\ref{fig:model-predictions} breaks down each models extracts, showing that GPT-5 is leading both in total volume and in unique coverage. Llama-4-17B and GPT-4o Mini also provide sizable unique segments, whereas Llama-3.3-70B and especially GPT-4o are more overlapping. Looking at the same stats for the IPBES dataset extractions, we found that almost 50\%  of the extracted statements (913) are shared across the LLMs. and 843 uniquely identified.

Figure 2 compares normalized performance across five knowledge gap categories \cite{boguslav2023creating} and shows near concentric contours, where GPT-5 traces the outermost polygon on every axis, followed by Llama-models, with GPT-4o Mini innermost. This consistent ordering across axes indicates shared ability in retrieving certain types of gaps.


\begin{figure}[t]
    \centering
    \includegraphics[width=0.65\linewidth]{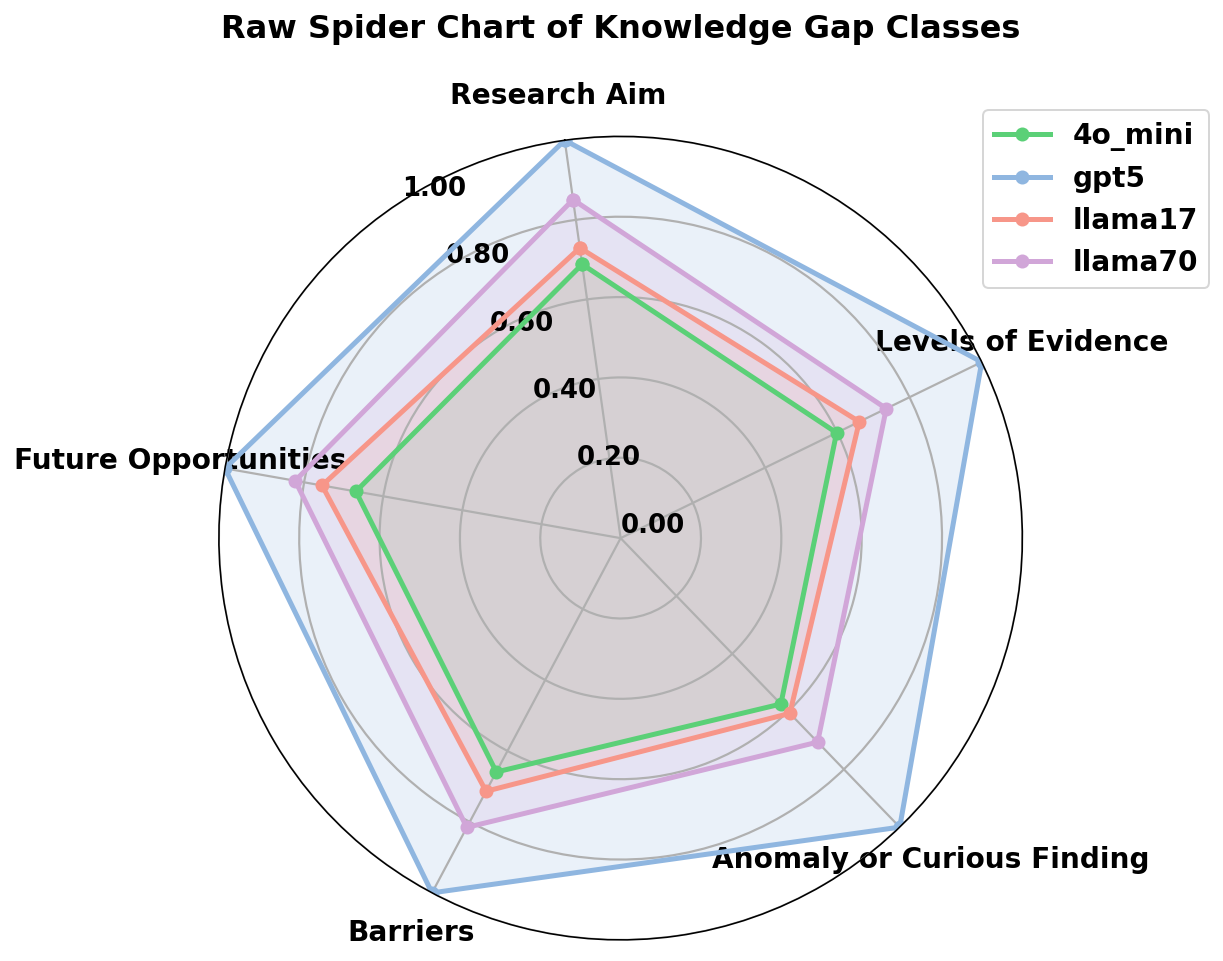}
    \caption{Model performance across knowledge-gap categories (Research Aim, Levels of Evidence, Anomaly/Curious Findings, Barriers, and Future Opportunities). The chart shows relative strengths of GPT-5, GPT-4o Mini, Llama-3.3-70B, and Llama-4-17B normalized.}
    \label{fig:gap-classes}
\end{figure}

\paragraph{Implicit Gap Conclusion from Paragraphs}

Table 5 reports accuracy on the paragraph-level. GPT-5 achieved the strongest performance, followed closely by GPT-4o and GPT-4o Mini, while mid-sized open-weight models such as Llama-3.3-70B showed competitive but lower alignment. Smaller models (e.g., Llama-3.1-8B, Gemma-2-9B) struggled to generalize. These results highlight that large LLMs can reliably conclude implicit claims when provided with sufficient guidance. 

A crucial factor in enabling this performance was the use of \emph{in-context 3-shot prompting}, rather than zero-shot extraction as we used to do with explicit extractions. When testing the LLMs with 0-shot learning similar to the explicit gaps extractions, even GPT-5’s outputs degenerated into vague restatements or unsupported speculations, resulting in sharp drops in accuracy. Thus, the design of the prompting strategy was essential: it provided not just examples (3-shot) but also a template that translated raw generative ability into precise, valuable future-direction predictions. 

We also found that the LLMs bucketed 10\% to 24\% of the correct future direction claims as less probable. While this is still a small fraction of bucketed predictions, it raises the question of what other factors can contribute to considering certain open questions more important than others such as relevance to laboratory and technical limitations.

\vspace{-3mm}
\begin{table}[ht]
\centering
\small                             
\setlength{\tabcolsep}{1pt}        
\renewcommand{\arraystretch}{1.05} 
\begin{tabular}{l*{7}{c}}
\toprule
 & \textbf{\rothead{GPT-5}} & \rothead{GPT-4o} & \rothead{GPT-4o mini} & \rothead{Gemma-2-9B} & \rothead{llama-3.3-70} & \rothead{llama-4-17B} & \rothead{llama-3.1-8b} \\
\midrule
FD Count & \textbf{179} & 171 & 171 & 48 & 163 & 135 & 65 \\
Accu. (\%)        & \textbf{84.43} & 80.66 & 80.66 & 22.64 & 77.89 & 63.68 & 30.67 \\
\bottomrule
\end{tabular}
\caption{LLMs accuracy in concluding future direction claims from paragraph dataset.}
\end{table}

\vspace{-9mm}
\paragraph{Implicit Gap Conclusion from Full Articles}Results indicate strong effectiveness: 83.3\% of participants agreed the model’s identified knowledge gaps were factually true, demonstrating GPT-4o’s accuracy. Regarding whether these questions remain open, 56\% fully agreed and 25.9\% partially agreed. Among the latter, 67\% believed that addressing the gaps might still significantly advance the field, underscoring their potential impact. Implementation was harder: only 65\% of proposed future directions were deemed valid, while 35\% were judged invalid mainly for feasibility reasons (technological limits, budget constraints, or relevance to other research groups). These findings highlight practical constraints on applying LLM-generated solutions and suggest that, although GPT-4o can effectively identify gaps, turning suggestions into actionable research requires attention to context and resources.

\vspace{-3mm}
\section{Conclusion}
\vspace{-3mm}
Our results prove that LLMs can be systematically identify explicit and implicit knowledge gaps in biomedical literature with high accuracies. We believe that using a mixture of LLMs instead of relying only on the top performing ones can provide a robust gap recommender systems. Such system cab identify high-impact open questions and align them with feasible next steps, thereby, focusing human effort and funding to accelerate scientific discovery.

\section*{Acknowledgments and Disclosure of Funding}
The authors gratefully acknowledge the support that made this research possible through the Chan Zuckerberg Initiative (CZI), Award ID AWD105712, for the *Rare as One Ignorome* project.

\end{document}